\documentclass[fleqn,10pt]{wlscirep}

\usepackage[utf8]{inputenc}
\usepackage[T1]{fontenc}
\usepackage{lmodern}
\usepackage{slantsc}
\usepackage{caption}
\usepackage{xspace}
\usepackage{xcolor}
\usepackage{enumitem}
\definecolor{myred}{rgb}{0.8,0,0}

\newcommand{\imagine}{\textsc{imagine}\xspace}

\newcommand{\ai}{AI\xspace}
\newcommand{\rl}{RL\xspace}

\newcommand{\nlp}{NLP\xspace}
\newcommand{\hrl}{HRL\xspace}

\newcommand{\llms}{LLMs\xspace}
\newcommand{\decstr}{\textsc{decstr}\xspace}
\newcommand{\ie}{i.e.\,}
\newcommand{\eg}{e.g.\,}

\newcommand{\todo}[1]{}
\renewcommand{\todo}[1]{\textcolor{myred}{\textbf{Todo: }\textit{#1}}}

\title{Language and Culture Internalisation for Human-Like Autotelic AI}

\author[1,*,+]{Cédric Colas}
\author[1,2,+]{Tristan Karch}
\author[1,3]{Clément Moulin-Frier}
\author[1,3,4]{Pierre-Yves Oudeyer}
\affil[1]{INRIA Bordeaux -- Sud-Ouest, Talence, France}
\affil[2]{Université de Bordeaux, Talence, France}
\affil[3]{ENSTA Paris, Palaiseau, France}
\affil[4]{Microsoft Research, Montreal, Canada}
\affil[*]{cedric.colas@inria.fr}
\affil[+]{equal contribution}


\begin{abstract} 
Building autonomous agents able to grow open-ended repertoires of skills across their lives is a fundamental goal of artificial intelligence (\ai). A promising developmental approach recommends the design of intrinsically motivated agents that learn new skills by generating and pursuing their own goals\,---\,\textit{autotelic agents}. But despite recent progress, existing algorithms still show serious limitations in terms of goal diversity, exploration, generalisation or skill composition. 
This perspective calls for the immersion of autotelic agents into \textit{rich socio-cultural worlds}, an immensely important attribute of our environment that shapes human cognition but is mostly omitted in modern \ai. 
Inspired by the seminal work of Vygotsky, we propose \textit{Vygotskian autotelic agents}\,---\,agents able to internalise their interactions with others and turn them into \textit{cognitive tools}. We focus on language and show how its structure and informational content may support the development of new cognitive functions in artificial agents as it does in humans.
We justify the approach by uncovering several examples of new artificial cognitive functions emerging from interactions between language and embodiment in recent works at the intersection of deep reinforcement learning and natural language processing. Looking forward, we highlight future opportunities and challenges for Vygotskian Autotelic AI research, including the use of language models as cultural models supporting artificial cognitive development. 

\end{abstract}

\begin{document}

\flushbottom
\maketitle

\thispagestyle{empty}

\section*{Introduction}
Humans are remarkable examples of lifelong open-ended learners. They learn to recognise objects and crawl as infants, learn to ask questions and interact with peers as toddlers, learn to master engineering, science, or arts as adults. A fundamental goal of artificial intelligence (\ai) is to build autonomous agents capable of growing such open-ended repertoires of skills. 

\begin{figure}[!ht]
   \centering
   \captionsetup{width=.85\linewidth}
   \includegraphics[width=\linewidth]{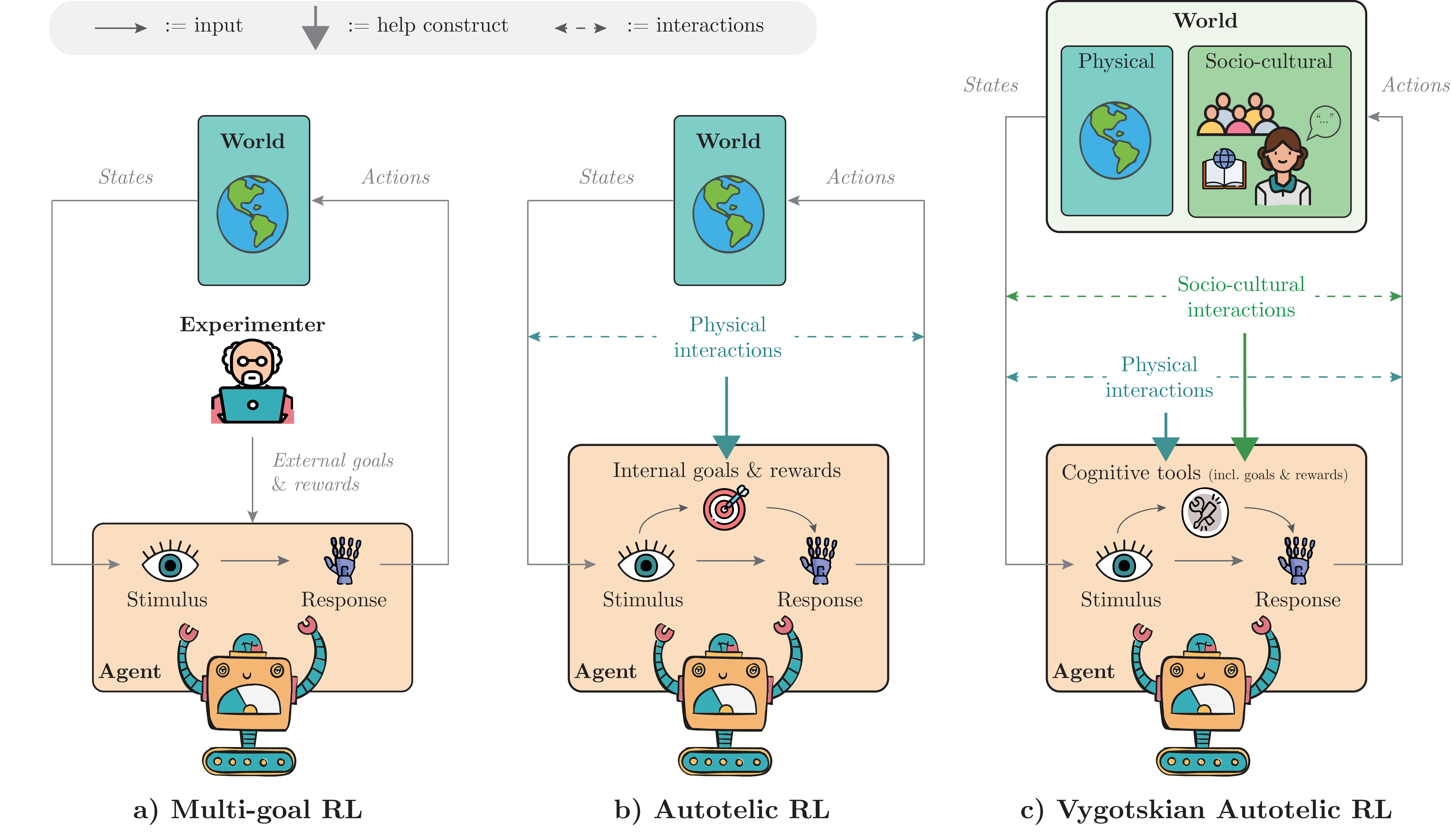}
    \caption{\small From multi-goal RL to autotelic RL to Vygotskian autotelic RL. \rl defines an agent experiencing the state of the world as stimuli and acting on that world via actions. Multi-goal RL (a): goals and associated rewards come from pre-engineered functions and are perceived as sensory stimuli by the agent. Autotelic RL (b): agents build internal goal representations from interactions between their intrinsic motivations and their physical experience (Piagetian view). Vygotskian autotelic RL (c): agents internalise physical and socio-cultural interactions into \textit{cognitive tools}. Here, \textit{cognitive tools} refer to any self-generated representation that mediates stimulus and actions: self-generated goals, explanations, descriptions, attentional biases, visual aids, mnemonic tricks, etc. }
    \label{fig:rl_arl_varl}
\end{figure}

\textit{Reinforcement learning} (\rl) offers a mathematical framework to formalise and tackle skill learning problems. For an embodied and situated \rl agent, \textit{learning a skill} (\eg playing chess) is about learning to act so as to maximise future \textit{rewards} measuring progression in that skill (\eg +1 for winning a game, -1 for losing it).\cite{sutton_introduction_1998} Extensions based on modern deep learning methods (deep \rl) have recently made the headlines by solving a wealth of complex problems: beating chess and go world champions,\cite{silver2016mastering} controlling stratospheric balloons,\cite{bellemare2020autonomous} or maintaining plasma in fusion reactors.\cite{degrave2022magnetic} But human chess world champions can also run, cook, draw a cat, or make a friend laugh. Humans are proficient in a wide diversity of tasks, most of which they just invent for themselves. In its standard form, the \rl framework considers a single predefined reward function and, thus, must be extended (see Figure~\ref{fig:rl_arl_varl}, a).

\vspace{.4cm}

\vspace{.4cm}

Piaget, a pioneer of developmental psychology, demonstrated children's ability to set their own goals, shape their own learning trajectories, and develop in symbiosis with their physical environment.\cite{piaget1952origins} This vision strongly influenced \ai research.\cite{dautenhahn_studying_1999} Indeed, if each skill is the association of a goal (\eg ``be a good chess player'') and a policy to reach it, then \textit{open-ended skill discovery} presupposes the ability to invent and select one's own goals and reward functions so as to progressively build repertoires of skills. \textit{Autotelic \textsl{\rl}}\,---\,from the Greek \textit{auto} (self) and \textit{telos} (goal)\,---\,extends the \rl framework to build such agents.\cite{colas2021intrinsically} It integrates two extensions of the standard \rl framework: the ability to consider multiple goals in parallel (\textit{multi-goal \textsl{\rl}}) and the ability to represent and select one's own goals. Although the first extension is straightforward,\cite{schaul_universal_2015} the second requires another ingredient inspired by the study of human learning: \textit{intrinsic motivations}. 

Most of the human time is spent on activities that do not seem to satisfy any utilitarian end; think about children playing, or adults watching movies. Psychologists argue that such exploratory behaviours are powered by \textit{intrinsic motivations} (IMs), a set of brain processes driving us to experience situations for the mere pleasure of experiencing novelty, surprise, or learning progress.\cite{berlyne_curiosity_1966,  kidd_psychology_2015, gottlieb_towards_2018} Similar processes can be coded into artificial agents to foster spontaneous exploratory capabilities:\cite{schmidhuber_curious_1991, barto_intrinsic_2005, oudeyer_intrinsic_2007} \textit{knowledge-based IMs} drive agents to experience parts of the environment to improve their internal models of the world and \textit{competence-based IMs} let them improve their mastery of self-generated goals.\cite{oudeyer_what_2009} Following the Piagetian tradition, autotelic agents see their goal representation and selection mechanisms emerge from interactions between IMs and their experience of the physical world (see Figure~\ref{fig:rl_arl_varl}, b).\cite{colas2021intrinsically}


In practice, current autotelic \rl implementations still lack human-like open-endedness. Their goal representations end up very concrete and mostly consist in reaching target stimuli (\eg matching their visual input with a particular target).\cite{colas2021intrinsically} This contrasts with the wide diversity and abstraction of goals targeted by humans. The generated goals often belong to the distribution of previously experienced effects, which drastically limits the ability of autotelic agents to represent \textit{creative goals}; thus to explore and undergo an open-ended discovery process.\cite{colas_language_2020} Besides goal imagination, \rl algorithms still lack human-like capacities in terms of generalisation, skill composition, abstraction, or sample efficiency.\cite{witty2021measuring, shanahan2022abstraction} 

In complement of Piaget's vision, we propose to take inspiration from the work of the developmentalist Vygotsky,\cite{vygotsky_thought_1934} further extended by many others in psychology,\cite{berk_why_1994,lupyan_what_2012,gentner_analogy_2017} linguistics,\cite{whorf_language_1956,rumelhart_sequential_1986,lakoff2008metaphors} and philosophy.\cite{hesse1988cognitive, dennett_consciousness_1993, carruthers_magic_1998, carruthers_modularity_2002} This tradition puts socio-cultural situatedness first.\cite{vygotsky_thought_1934, tomasello_cultural_1999, tomasello_understanding_2005, brewer2014addressing} The most impressive human cognitive abilities may first appear as social processes, before we \textit{internalise} them within us and turn them into cognitive functions during our cognitive development.\cite{vygotsky_thought_1934} 

We advocate for a \textit{Vygotskian Autotelic AI}. Autotelic agents must be immersed into our rich socio-cultural world; internalise social interactions and mesh them with their cognitive development (see Figure~\ref{fig:rl_arl_varl}, c). Just like they do for humans, language and culture will help agents represent and generate more diverse and abstract goals; to imagine new goals beyond their past experience. Because they will develop at our contact, bathed in our cultures, Vygotskian autotelic agents will learn about our cultural norms, values, customs, interests, and thought processes; all of which would be impossible to learn in social isolation. Just like humans, machines will use language to develop higher cognitive functions like abstraction, generalisation, or imagination.\cite{carruthers_language_1998, gentner_analogy_2017, dove_language_2018}  

Recent advances in \ai make our proposition particularly timely. These past years have indeed seen a revolution in natural language processing (\nlp), the set of tools designed to analyse and generate language. Generative models of language trained on gigantic amounts of text such as GPT-3, PaLM, or Flamingo can now produce high-quality language,\cite{brown2020language,chowdhery2022palm} handle multimodal inputs,\cite{radford2021learning,ramesh2022hierarchical,alayrac2022flamingo}, solve complex arithmetic tasks involving multi-step reasoning\cite{creswell2022selection}, explain jokes and memes\cite{alayrac2022flamingo}, capture common sense\cite{west_symbolic_2021} or cultural artefacts.\cite{hershcovich_challenges_2022,arora2022probing} 
The ongoing convergence of autotelic \rl and \nlp will offer a wealth of opportunities as autotelic agents learn to interact with us, learn from us, and teach us back. This motivates the elaboration of a theoretical framework to understand recent linguistic \rl developments and point towards future challenges in the quest for open-ended skill discovery.

This perspective extends previous calls to leverage Vygotsky's insights for a more socially-situated cognitive robotics.\cite{dautenhahn_studying_1999, zlatev_epigenesis_2001, lindblom_social_2003,mirolli_towards_2011} Zlatev discussed interactions between social-situatedness and epigenetic development,\cite{zlatev_epigenesis_2001} Dautenhahn and Billard drew the parallel between \ai and the Piagetian vs.\, Vygotskian views,\cite{dautenhahn_studying_1999} while Mirolli and Parisi, as well as Cangelosi et al., reviewed the first successful auxiliary uses of language for decision-making.\cite{mirolli_towards_2011, cangelosi2010integration} In the last decade however, the \ai community seems to have lost track of these insights. Today, we update these arguments in light of recent \ai advances and re-frame the Vygotskian perspective within the autotelic \rl framework, with a specific focus on language-based socio-cultural interactions. As a result, we will not discuss non-embodied multi-modal supervised techniques\cite{radford2021learning, ramesh2022hierarchical, alayrac2022flamingo} or non-linguistic autotelic \rl.\cite{schaul_universal_2015} We will not cover advances in the  computational modelling of non-linguistic socio-cultural interactions (social \rl).\cite{jaques2019social} Although future Vygotskian autotelic agents must incorporate social \rl mechanisms, current approaches do not consider autotelic agents able to set their own goals and, as such, do not tackle the open-ended skill discovery problem (see a complete discussion in a related paper\cite{sigaud_towards_2021}).

The next section sets the background and discusses the interaction between language and thought in humans by building on the work of psychologists and philosophers (Section~\ref{sec:4views}). Next, we dive into the two key elements of a Vygotskian autotelic \ai: 1)~the ability to exploit linguistic information to support the development of cognitive functions (Section~\ref{sec:extraction}); 2)~the ability to internalise linguistic interactions within the agent to further develop cognitive functions and autonomy (Section~\ref{sec:production}). Finally, Section~\ref{sec:challenges} identifies open challenges for future research.


\section{A Vygotskian Perspective on Language and Thought in Humans}
\label{sec:4views}
The human ability to generate new ideas is the source of our incredible success in the animal kingdom. But this ability did not appear with the first \textit{homo sapiens} 130,000 years ago. Indeed, the oldest imaginative artefacts such as figurative arts, elaborate burials, or the first dwellings only date back to 70,000 years ago.\cite{harari_sapiens_2014, vyshedskiy_language_2019} This is thought to coincide with the apparition of \textit{recursive language}.\cite{goldberg1999emergence, vyshedskiy_language_2019, hoffmann_construction_2020} Which of these appeared first? Creativity or recursive language? Or did they mutually bootstrap?

Extreme views on the topic either characterise language as a pure communicative device to convey our inner thoughts (strong communicative thesis)\cite{chomsky_syntactic_1957,fodor1975language} or, on the other extreme, argue that only language can be the vehicle of our thoughts (strong cognitive thesis)\cite{wittgenstein1953philosophical, mcdowell1996mind} As often, the truth seems to lie in between. Animal and preverbal infants demonstrate complex cognition,\cite{sperber1995causal, allen1999species} but language does impact the way we perceive,\cite{waxman_words_1995, yoshida_sound_2003} represent concepts,\cite{lakoff2008metaphors} or conduct compositional and relational thinking,\cite{gentner2002relational, gentner_analogy_2017, vyshedskiy_language_2019}. Thus, language seems to be at least \textit{required} to develop some of our cognitive processes (requirement thesis), and might still be the vehicle of \textit{some} of our thoughts (constitutive thesis).\cite{carruthers_language_1998} Interested readers can find a thorough overview of this debate in \textit{Language and Thought} by Carruthers and Boucher.\cite{carruthers_language_1998}

If language is required to develop some of our higher cognitive functions, then autotelic artificial agents should use it as well. But how does that work? What is so special about language? 
Let us start with \textit{words}, which some called \textit{invitations to form categories}.\cite{waxman_words_1995} Hearing the same word in a variety of contexts invites humans to compare situations, find similarities, differences and build symbolic representations of agents, objects and their attributes. With words, the continuous world can be simplified and structured into mental entities at multiple levels of abstraction. 

The recursivity and partial compositionality of language allow us to readily understand the meaning of sentences we never heard before by generalising from known words and syntactic structures. On the flip side, it also supports \textit{linguistic productivity},\cite{chomsky_syntactic_1957} the ability to generate new sentences\,---\,thus new ideas\,---\,in an open-ended way. 
Relational structures such as comparisons and metaphors facilitate our relational thinking,\cite{gentner2002relational, gentner_analogy_2017} condition our ability to compose mental images,\cite{vyshedskiy_language_2019} and support our understanding of abstract concepts such as emotions, politics or scientific theories.\cite{hesse1988cognitive,lakoff2008metaphors} 

Finally, language is a cultural artefact inherited from previous generations and shared with others. It supports our cultural evolution and allows humans to efficiently transfer knowledge and practices across people and generations\cite{henrich2003evolution, morgan_experimental_2015, chopra2019first}\,---\,a process known as the \textit{cultural rachet}.\cite{tomasello_cultural_1999} Through shared cultural artefacts such as narratives, we learn to share common values, customs and social norms, we learn how to navigate the world, what to attend to, how to think, and what to expect from others.\cite{bruner1990acts} This cultural knowledge is readily accessible to children as they enter societies via social interactions and formal education. Learning a language further extends access to cultural artefacts such as books, movies, or the Internet. These act as a thousand virtual social partners to learn from. 

We now understand why language is so special. Let us focus on how it can shape cognitive development in humans and machines. Dennett, a proponent of the requirement thesis, suggests that linguistic exposition alone can lead to a fundamental cognitive reorganisation of the human brain.\cite{dennett_consciousness_1993} He compares it to the installation of a serial virtual machine on humans' massively parallel processing brains. As a result, a slight change in our computational hardware compared to our primate relatives could open the possibility for a language-driven cognitive software reprogramming which, in turn, could trigger the learning and cultural evolution of higher cognitive capacities.
Carruthers, a proponent of the constitutive thesis, suggests that language may have evolved as a separate module to exchange inner representations with our peers (naive physics, theory of mind). This would require connections between linguistic and non-linguistic modules to allow conversions between inner representations and linguistic inputs/outputs. In a similar way that humans can trigger imagined visual representations via top-down connections in their visual cortex, top-down activations of the linguistic module would create \textit{inner speech}. This hallucinated speech, when broadcast to other modules, would implement \textit{thinking in language}.\cite{carruthers1998thinking} Clark advances yet another possibility, the \textit{supra-communicative view}. Here, language does not transform the way the brain makes computations and is not the vehicle of thoughts. Instead, language complements our standard computation activities by ``\textit{re-shaping the computational spaces},'' turning problems that would be out of reach into problems our pattern-matching brains can solve.\cite{carruthers_magic_1998} In that sense, language is a \textit{cognitive tool} that enhances our cognitive abilities without altering them per se. 

Vygotsky's theory brings a complementary argument to this debate. Caretakers naturally scaffold the learning experiences of children, tailoring them to their current objectives and capacities. Through encouragement, attention guidance, explanations or plan suggestions, they provide cognitive aids to children in the form of interpersonal social processes.\cite{vygotsky_thought_1934} In this \textit{zone of proximal development}, as Vygotsky coined it, children can benefit from these social interactions to achieve more than they could alone. In these moments, children \textit{internalise} linguistic and social aids and progressively turn these interpersonal processes into intrapersonal \textit{psychological tools}.\cite{vygotsky_thought_1934} This essentially consists in building internal models of social partners such that learners can self-generate contextual guidance in the absence of an external one. Social speech is internalised into private speech (an outer speech of children for themselves), which, as it develops, becomes more goal-oriented and provides cognitive aids of the type caretakers would provide.\cite{vygotsky_thought_1934,berk_why_1994} It progressively becomes more efficient and abbreviated, less vocalised, until it is entirely internalised by the child and becomes \textit{inner speech}. 

\begin{figure}[!ht]
    \centering
    \captionsetup{width=.85\linewidth}
    \includegraphics[width=1\linewidth]{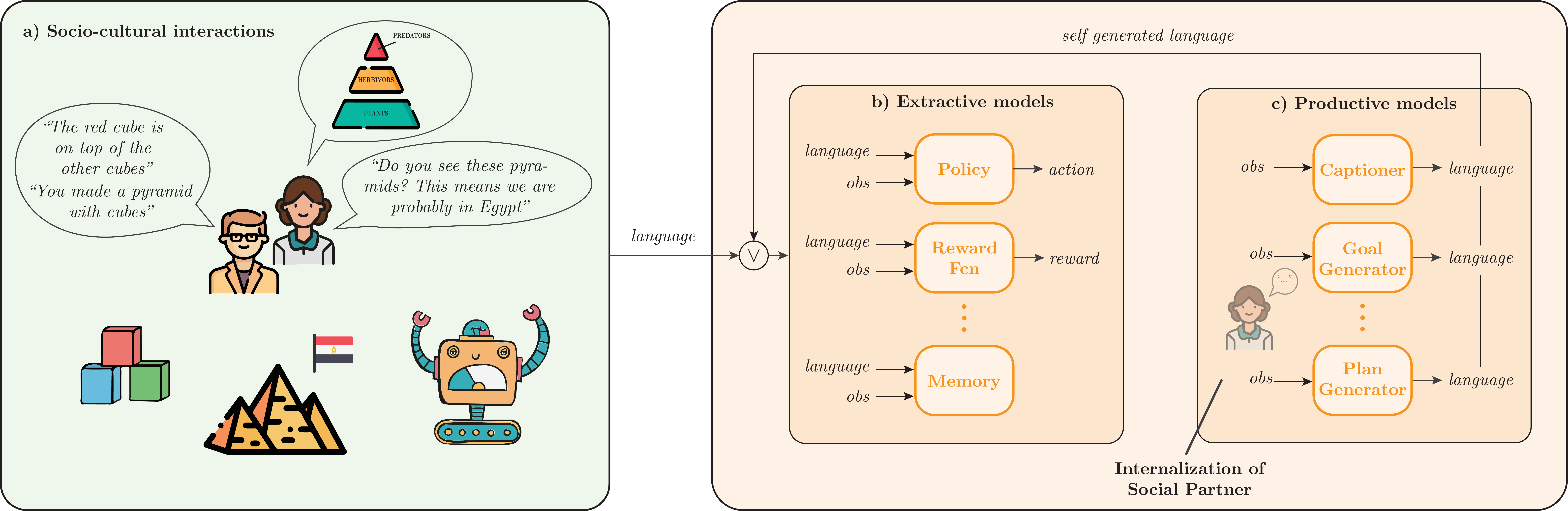}
    \caption{\small The three components of Vygotskian autotelic agents: socio-cultural interactions, linguistic extraction and internalised linguistic production. Vygotskian autotelic agents are immersed into rich socio-cultural worlds where they experience a variety of linguistic feedback including descriptions, explanations, or metaphors (a). They can exploit information from linguistic structures and content by conditioning their internal modules on this feedback (b, extractive models). Finally, they learn to internalise social interactions by training productive models of language to generate feedback similar to the one they receive from others (c, productive models). This offers agents the autonomy to build their own cognitive tools, bootstrapped by socio-cultural language.}
    \label{fig:extractive_productive}
\end{figure}

This section showed why language is so important and might just be required for the development of our highest cognitive functions. From the arguments above, we identify two key elements, see Figure~\ref{fig:extractive_productive}. First, we need autotelic agents to exploit the information contained within linguistic structures and contents. Exposed to language, agents will reorganise their internal representations for better abstraction, generalisation and better alignment with human values, norms and customs (Dennett's thesis). Second, we need autotelic agents to internalise social interactive processes, \ie to \textit{model social partners within themselves} (Vygotsky's internalisation). Social processes, turned into intrapersonal cognitive processes will orient the agent's focus, help it decompose tasks or imagine goals. This inner speech generation can serve as a common currency between other modules (\eg perception, motor control, goal generation) in line with Carruther's view and will help agents project problems onto linguistic spaces where they might be easier to solve (Clark's view). In the next two sections (\ref{sec:extraction}--\ref{sec:production}), we discuss these two key elements in detail and re-frame recent works at the intersection of \rl and language in their light. 

\vspace{.4cm}
\vspace{.4cm}

\section{Exploiting Linguistic Structure and Content}
\label{sec:extraction}

Vocabulary, syntax and narratives contain both powerful computational tools for thinking and essential cultural knowledge about the world. In line with Dennett's thesis, recent \ai research shows that simple language exposure can trigger the rewiring of \rl agents' inner processes and, in this way, lead to the emergence of new cognitive abilities.

\textbf{Learning to abstract and generalise.} Exposure to linguistic labels is known to facilitate category learning in humans,\cite{waxman_words_1995, yoshida_sound_2003} but also in machines.\cite{lupyan_carving_2005} As Mirolli and Parisi defend, the repeated occurrence of a linguistic label (\textit{red} in their example) leads to the conflation of internal representations associated with that label (red things) which, in turn, facilitates further classifications based on the linguistic attributes.\cite{mirolli_towards_2011}

A similar effect can be spotted in \rl: agents exposed to aligned instructions and trajectories see the internal representations contained within their action policy reshaped by language. This occurs within a neural-network-based policy: a function conditioned on the agent's instruction that maps the current state of the world to the next action. If repeatedly asked to \textit{grasp red objects}, the policy will focus on objects' colours to facilitate action selection.\cite{hill_emergent_2019, colas_language_2020} The concept of \textit{red}\,---\,originally a cultural abstraction over a continuous space of colours\,---\,gets progressively internalised within the representations of the agent via a combination of language exposure and decision-making.

Exposed to a diversity of instructions, agents thus develop \textit{abstraction} capabilities. They learn to reach and make sense of abstract relational goals ``\textit{sort objects by size},''\cite{jiang_language_2019} ``\textit{put the cylinder in the drawer},''\cite{lynch_grounding_2020} sequential goals ``\textit{open the yellow door after opening a red door},''\cite{chevalier-boisvert_baby-ai_2019} or even learning goals ``\textit{is the ghargh edible?}.''\cite{yuan_interactive_2019} Whereas handling abstract goals used to require engineers to hard-code specific goal representations and reward functions within the agent,\cite{curious,team2021open} linguistic goals offer abstraction via simple linguistic interactions.\cite{bahdanau_learning_2019, colas_language_2020} 

With new abstractions distilled within their representations, agents can become better \textit{explorers}. Searching for novelty in a space of abstract linguistic descriptions of the world is indeed more efficient than searching for novelty in low-level sensorimotor spaces where novelty could be trivially triggered by leaves moving in the wind or TV noise.\cite{tam2022semantic, Mu2022ImprovingIE}

Abstraction also leads to the emergence of another cognitive ability called \textit{systematic generalisation}. Language-instructed agents demonstrate the ability to generalise to new instructions obtained by systematic recombinations of those they were trained on.\cite{hill_emergent_2019} For instance, agents that learned to \textit{grasp blue objects} and \textit{put green objects on the table} can directly \textit{grasp green objects} and \textit{put blue objects on the table}.\cite{hermann_grounded_2017, chevalier-boisvert_baby-ai_2019, bahdanau_learning_2019, hill_emergent_2019, hill_human_2020, colas_language_2020, sharma2021skill, karch2021grounding} This ability can either be encoded in learning architecture through the use of modular networks (neuro-symbolic approaches), or emerge spontaneously in plain networks under the right environmental conditions.\cite{hill_emergent_2019} Although sometimes the world does not conform to strict linguistic compositionality, systematic generalisation still supports good priors\,---\,\eg \textit{feeding the cat} is not a strict transposition of \textit{feeding the plant}, but they still share similarities (bringing supplies to the cat/plant).\cite{colas_language_2020} 

\textbf{Learning to represent possible futures.} After being exposed to aligned trajectories and linguistic descriptions, agents can generate concrete examples of abstract descriptions. The \decstr approach, for example, trains a generative world model to sample from the distribution of possible future states matching a given abstract linguistic description.\cite{akakzia_grounding_2021} This simple mapping supports \textit{behavioural diversity}, the ability to represent different possible futures so as to select one to pursue. Recently, text-to-image generative models trained on pairs of images and compositional descriptions, have demonstrated impressive capacities to generate high-quality images from the most twisted descriptions humans can think of.\cite{ramesh_zero-shot_2021, ramesh2022hierarchical}
This ability to compose images creatively and to generate concrete visual instances of abstract descriptions directly emerges from the system's exposure to aligned language and images. Equipped with such systems, autotelic agents could develop new capacities for the visual imagination of possible futures. 

\textbf{Learning to decompose tasks.} Vygotsky and others discovered that children's use of private speech helps them increase self-control and is instrumental to their capacity to reason and solve hard tasks.\cite{vygotsky_thought_1934, berk_why_1994} The ability to formulate sentences like ``at the left of the blue wall,'' for instance, predicts spatial orientation capacities in such contexts, while interfering with adult's inner speech via speaking tasks seems to hinder theirs.\cite{hermer-vazquez_language_2001}

Language indeed contains cues about how to decompose tasks into sub-tasks, \ie how to \textit{generate good plans}. Although \textit{gharble} is a made-up word, \textit{fry the gharble} probably involves a preparation of the gharble (\eg peeling, cutting), some sort of oil and a frying pan.\cite{yuan_interactive_2019} \textit{Draw an octogon} contains the prefix \textit{octo}, so we should probably do something 8 times.\cite{wong_leveraging_2021} Recent \ai approaches leverage these regularities by training \textit{plan generators} from linguistic task descriptions.\cite{jiang_language_2019, chen_ask_2021, sharma2021skill, mirchandani2021ella, shridhar_alfworld_2021, wong_leveraging_2021} Among them, Wong et al.\, generate plan descriptions as an auxiliary task to train a drawing policy.\cite{wong_leveraging_2021} This auxiliary task, only used in the training phase, shapes the internal representation of the main policy which, they find, favours abstraction and generalisation in the main drawing task. 

Inspired by video games of the 80s such as \textit{Zork}, text-based environments define purely linguistic goals, actions and states.\cite{cote_textworld_2018, das_embodied_2018, yuan_interactive_2019} Training a policy in such environments can be seen as training a plan generator in a linguistic world model, \ie training an inner speech to generate good task decompositions. This idea was exploited in \textit{AlfWorld}, where a pre-trained plan generator is deployed in a physical environment to generate sub-goals for a low-level policy.\cite{shridhar_alfworld_2021} Here, the abstraction capabilities of language help the plan generator solve long-horizon tasks.

The above approaches echo the thesis of Dennett (Section~\ref{sec:4views}): the mere exposure to structured language, once internalised within internal modules (reward function, policy, world model) strongly shapes inner representations in new ways and supports new cognitive functions (abstraction, structured exploration, future states generation, compositional generalisation, task decomposition, etc).

\textbf{Learning from cultural artefacts.} Large language models (\llms) are trained on huge quantities of text scrapped from the internet: Wikipedia, forums, blogs, scientific articles, books, subtitles, etc.\cite{devlin2019bert,brown2020language} We can understand them as (imperfect) \textit{cultural models} containing information about our values, norms, customs, history and interests.\cite{hershcovich_challenges_2022,arora2022probing} This represents exciting opportunities for autotelic agents to learn about us, align with us, and better navigate our complex world. So far, researchers have leveraged the zero-shot capacities of \llms to decompose tasks into high-level plans,\cite{huang2022language} to score available actions,\cite{ahn2022doasican} or to suggest creative goals in Minecraft.\cite{fan2022minedojo} We believe that these works only scratch the surface of what these cultural models can do for open-ended skill discovery and will discuss more opportunities in the Challenge~\#3 of Section~\ref{sec:challenges}. 

\section{Internalisation of Language Production}
\label{sec:production}

Agents that internalise extractive models learn to exploit the information contained within linguistic vocabularies, structures and narratives. However, most of them require external linguistic inputs at test time and, thus, cannot be considered autonomous. Vygotskian autotelic agents reach autonomy by internalising \textit{productive models}; \ie by learning to generate their own linguistic inputs, their own \textit{inner speech} (see Figure~\ref{fig:extractive_productive}, c). 

Inner speech can be understood as a fully-formed language: descriptions, explanations or advice to be fed back to extractive models; to serve as a common currency between cognitive modules (fully-formed inner speech).\cite{zeng2022socratic} But it might also be understood as \textit{distributed representations} within productive models, \ie \textit{upstream} from fully-formed language (distributed inner speech). In the latter interpretation, linguistic production acts as an auxiliary task whose true purpose is to shape the agent's cognitive representations. Symbolic behaviours might indeed not require explicit symbolic representations but may emerge from distributed architectures trained on structured tasks, \eg involving linguistic predictions.\cite{mcclelland2010letting,santoro2021symbolic} 
In the literature, we found four types of productive models making use of either fully-formed or distributed inner speech: trajectory captioners, plan generators, explanation generators and goal generators. 

\textbf{Trajectory captioners.} Trajectory captioners are trained on instructive or descriptive feedback to generate valid descriptions of scenes or trajectories.\cite{cideron_higher_2020, zhou_inverse_2020, colas_language_2020, nguyen2021interactive, yan2022intra} In line with Vygotsky's theory, these agents internalise models of descriptive social partners. They generate an inner speech describing their ongoing behaviours just like a caretaker would. Used as an auxiliary task (distributed inner speech), description generation shapes an agent's representations such that it generalises better to new tasks.\cite{yan2022intra} With fully-formed inner speech, agents can generate new multi-modal data autonomously, and learn from past experience via \textit{hindsight learning},\cite{andrychowicz_hindsight_2017} \ie the reinterpretation of their trajectory as a valid behaviour to achieve the trajectory's description.\cite{zhou_inverse_2020, colas_language_2020, nguyen2021interactive}

\textbf{Plan generators.} Plan generators are both extractive and productive. Following the formalism of hierarchical \rl (\hrl), plan generators are implemented by a \textit{high-level policy} generating linguistic sub-goals to a low-level policy (executioner).\cite{dayan_feudal_1993, sutton_between_1999} Linguistic sub-goals are a form of inner speech that facilitates decision-making at lower temporal resolution by providing abstract, human-interpretable actions, which themselves favour systematic generalisation for the low-level policy (see Section~\ref{sec:extraction}).\cite{jiang_language_2019, chen_ask_2021, shridhar_alfworld_2021} Here, agents internalise linguistic production to autonomously generate further guidance for themselves in fully-formed language (task decompositions). 

\textbf{Explanation generators.} Vygotskian agents can generate \textit{explanations}. Using the generation of explanations as an auxiliary task (distributed inner speech) was indeed shown to support causal and relational learning in complex disembodied \textit{odd one out} tasks.\cite{lampinen_tell_2021} 

\textbf{Goal generators.} Some forms of creativity appear easier in linguistic spaces because swapping words, compositing new sentences, and generating metaphors are all easier in the language space than in sensorimotor spaces. The \imagine approach leverages this idea to support \textit{creative goal imagination}.\cite{colas_language_2020} While previous methods were limited to generating goals within the distribution of past experience (\eg with generative models of states\cite{nair2018visual}), \imagine invents out-of-distribution goals by combining descriptions of past goals. These manipulations occur in linguistic spaces directly and are thus \textit{linguistic thoughts}; fully-formed inner speech (Carruthers' view). The problem of goal imagination, difficult to solve in sensorimotor space, is projected onto the linguistic space, solved there, and projected back to sensorimotor space (Clark's view). This, in turn, powers additional cognitive abilities: a more structured exploration oriented towards objects interactions, and an enhanced systematic generalisation caused by widening the set of goals the agent can train on.\cite{colas_language_2020}

By internalising linguistic production, \imagine generates goals that are both \textit{novel} (new sentences) and \textit{appropriate} (they respect linguistic regularities, both structures, and contents).\cite{runco_standard_2012} Social descriptions focus on objects, object attributes, and interactions with these objects. Imagined goals obtained by recompositions of social ones share the same attentional and conceptual biases, \eg by reusing semantic categories of a particular culture. Thus, cultural biases are implicitly transmitted to the agent, which forms goal representations and biases goal selection following cultural constraints.\cite{colas_language_2020}



\section{Open Challenges}
\label{sec:challenges}
We identify four main challenges for future research. 

\textbf{Challenge \#1: Immersing autotelic agents in rich socio-cultural worlds.}
Vygotskian autotelic agents must be immersed into rich socio-cultural worlds close to ours. To this end, we must both enrich the socio-cultural aspects of learning environments and augment the agents' capacities for interactivity and teachability.

Socio-cultural learning environments must allow multimodal perceptions to account for all aspects of cultural transmissions. Cultural interactions may indeed involve non-verbal aspects, including motor, perceptual or emotional dimensions. In addition, autotelic agents must interact with other agents and with humans. Scaling the socio-cultural richness of these worlds may thus require the involvement of the video game industry\,---\,expert in the design of complex and realistic multimodal worlds. Human-in-the-loop research will also be required to let humans enter these virtual worlds, for instance via virtual reality technology. 

Vygotskian autotelic agents must be more interactive and teachable. In a recent paper, Sigaud and colleagues discuss this challenge through a detailed analysis of children's learning abilities and teacher-child interactions.\cite{sigaud_towards_2021} They present a checklist of properties that future Vygotskian autotelic agents must demonstrate to be considered \textit{teachable}. To interact with humans, Vygotskian autotelic agents will need to target goals in multiple modalities (\eg linguistic, perceptual, emotional) with various levels of abstraction.\cite{colas2021intrinsically}. By doing so, they could leverage cross-domain hindsight learning: using experience collected while aiming at a goal to learn about other goals in other domains.\cite{curious} 

\textbf{Challenge \#2: Enabling artificial mental life with a more systematic internalised language production.}
Only a few approaches internalise language production within agents (Section~\ref{sec:production}). So far limited to a few use cases, language production should concern every possible linguistic feedback agents could receive: instructions, corrections, advice, plan suggestions, explanations, or cultural artefacts. This inner speech may be the embryo of an \textit{artificial mental life}. Looping back to the constitutive thesis of Carruthers presented in Section~\ref{sec:4views}, inner speech acts as a common currency for inner modules to exchange information (see a recent implementation of this idea\cite{zeng2022socratic}). Combined with world models, inner speech could trigger the simulation of perceptual experiences (images, sounds), sensorimotor trajectories, the imagination of possible futures or the recall of memories. Observing these internal simulations, agents may produce new behaviours and new inner speech. These inner loops support a mental life that could help agents reason; trigger memories or mnemonic representations acting as cognitive aids. This account is compatible with the embodied cognition hypothesis, which suggests that thinking and sensorimotor simulation are one and the same. Here, language brings another set of abstract, compositional and recursive inputs and outputs for simulations. As argued by Dove,\cite{dove_language_2018} simulating such abstract linguistic structures (words, metaphors, etc) might offer us the capacity to reason abstractly.

\textbf{Challenge \#3: Building editable and shareable cultural models with aligned LLMs.} 
Vygotskian autotelic agents could leverage \llms as (imperfect) cultural models in order to learn about foundational human concepts, causality, folk psychology, politeness, ethics and all the physical or cultural information that are the subject of everyday stories: fiction, news, or even simple narratives parents use to explain everyday things to their children. LLMs could be seen as proxies to human cultures, offering both opportunities and challenges.

With current \llms, agents could extract common sense knowledge to generate creative goals to support exploration, or a succession of goals forming a natural curriculum for the agent based on its current abilities and descriptions of the world around it. Agents could use them to predict the outcomes of its actions given the context and use this to plan in abstract search spaces. Agents could also ask \llms for guidelines only when they cannot solve the problem themselves (active learning) and, more generally, to augment the world state with commonsense knowledge.

But letting autotelic agents rely on \llms also poses challenges. \llms are known to convey false information and harmful biases that may come directly from the data sources or be introduced through inadequate training processes.\cite{shah-etal-2020-predictive,pmlr-v139-liang21a,weidinger2021ethical,bender2021dangers} Autotelic agents relying on these models could demonstrate harmful behaviours and contribute to reinforcing stereotypes and inequalities. Ideally, autotelic agents equipped with \llms should align with the objectives and needs of a particular family of users, tasks, and contexts. This will require advances in bias mitigation strategies\cite{pmlr-v139-liang21a,bender2021dangers} and alignment methods to make \llms more reliable, trustworthy and moral. 

Cultural information, by definition, biases the search space. This can be useful when the search space is large (\eg knives are found in kitchens), but may also limit exploration and lead the learner to prematurely give up on promising avenues\cite{bonawitz2011Double} (\eg in astronomy, the cultural support for the geocentric model significantly delayed the acceptance of the heliocentric model). This is true for both humans and artificial agents. 

During their education, children are taught to think critically; for themselves. Autotelic agents should be taught in the same way. Because they are autonomous embodied machines, autotelic agents can conduct experiments in the world and empirically test the information they were provided. This physical embodiment is often described as the missing piece for \llms to truly understand the world.\cite{bisk2020experience} Just like human cultural narratives can be shifted by government, policies, advertising, activism, and pop culture, artificial cultural narratives should become more malleable. Autotelic agents must be given the possibility to steer, edit and extend their cultural models (\ie \llms) in light of their embodied experience; to share it and negotiate it with others; \ie to participate in a shared cultural evolution. Reversely, humans that train language/cultural models should pay great care to building and understanding the cultural input they provide, just as they pay close attention to raising their children. 

With a combination of \llms and true human-machine interactions, agents could learn some of the uniquely human social features described by Tomasello in his recent book\cite{Tomasello+2019}: cooperative thinking, moral identity or social norms. But the mere exposure to culture might not be sufficient to design truly autonomous social learners. This may require encoding certain mechanisms such as joint intentionality or other collaborative priors inside agents. These questions are the subject of the social \rl field.\cite{jaques2019social, velez2021learning}

\textbf{Challenge \#4: Pursuing long-term goals.} Current autotelic agents mostly pursue goals at the timescale of an episode. Humans, on the other hand, can pursue goals they can barely hope to achieve within their lifetime (\eg building an efficient fusion reactor). Because there is an infinity of potential goals and little time to explore them alone, autotelic agents may need cultural models to bias their selection of long-term goals towards more feasible, interesting, or valuable options\,---\,turning an individual exploration into a \textit{population-based exploration}.  Keeping long-term goals in mind will require improvements in architecture's memory systems, but might also benefit from language and culture. Indeed, verbalisation is known to increase humans' memory span \cite{elliott_multilab_2021} and writing let us set our goals in stone (from the post-it note reminding us to take the garbage out to the Ten Commandments). Young children progressively become future-oriented as they are taught to project themselves into the future through education, social interactions (\textit{what do you want to do when you grow up?}) and cultural metaphors (\eg the self-made man).\cite{atance2008future} If autotelic agents will need better hierarchical \rl algorithms to achieve long-term goals, they could also leverage cultural artefacts evolved for improved collaboration and long-term planning\,---\,think of roadmaps, organisation systems and
project management tools.\cite{carruthers_magic_1998} Because long-term goals are not immediately rewarding, human cultures supply shorter-term reward systems to track progress on these goals (good grades in the educational system, money and social recognition in professional careers)\,---\,a form of reward shaping. 

\section*{Conclusion}

This perspective builds on the autotelic \rl framework inspired by the Piagetian tradition to propose a complementary view motivated by Vygotsky's work: a \textit{Vygotskyan Autotelic AI}. It proposes to immerse artificial agents in rich socio-cultural worlds to allow them to transform their linguistic interactions into cognitive tools and become better learners. Recent works at the intersection of deep \rl and \nlp present promising first steps, but a lot remains to be done towards the design of more human-like autotelic agents. As research progresses on the training of cultural models, interactive capabilities and non-linguistic functions such as memory, future Vygotskian autotelic agents will learn to better interact with us and with our culture, exploit linguistic structures and content and finally internalise social interactions to serve as the basis of their future cognitive functions: abstraction, compositional thinking, generalisation or imagination. 

Vygotskian Autotelic \ai opens a new research program with exciting opportunities. The current \nlp revolution can be harnessed to design more interactive and teachable agents. As they interact with humans and their peers in rich socio-cultural worlds, Vygotskian autotelic agents might learn to leverage cultural and linguistic knowledge to bootstrap their cognitive development, and may eventually contribute back to our shared cultural evolution. This perspective uncovers an important challenge: how will we train and leverage cultural models to align with the values of human cultures so as to educate ethical and useful artificial agents?

Dennett used the software/hardware metaphor to describe the impact of language on human brains.\cite{dennett_consciousness_1993} Current \ai research mostly focuses on \textit{hardware} by asking \textit{how can we design better learning architectures and algorithms?} In complement, this paper suggests focusing on \textit{software} as well: \textit{How can we build the right socio-cultural bath which, combined with efficient hardware, may allow the emergence of a more human-like \ai}?




\section*{Acknowledgements}
We thank Olivier Sigaud for his helpful feedback. We also thank Andrew Lampinen and the anonymous reviewers for their detailed reviews and suggestions. C\'edric Colas and Tristan Karch are partially funded by the French Minist\`ere des Arm\'ees - Direction G\'en\'erale de l’Armement. Clément Moulin-Frier is partially funded by the Inria Exploratory action ORIGINS (\url{https://www.inria.fr/en/origins}) as well as the French National Research Agency (\url{https://anr.fr/}, project ECOCURL, Grant ANR-20-CE23-0006). 

\section*{Competing Interests Statement}
 The authors declare that they have no competing interests.


\end{document}